\documentclass[runningheads]{llncs}

\usepackage[mobile]{eccv}

\allowdisplaybreaks[1]

\usepackage{eccvabbrv}
\usepackage{multirow}
\usepackage{graphicx}
\usepackage{booktabs}
\usepackage{array}
\usepackage{wrapfig}
\usepackage{tabularx}
\usepackage[accsupp]{axessibility}  

\usepackage[pagebackref,breaklinks,colorlinks,citecolor=eccvblue]{hyperref}

\usepackage{orcidlink}
\usepackage{dashrule}

\begin{document}

\title{City-on-Web: Real-time Neural Rendering of Large-scale Scenes on the Web} 


\author{Kaiwen Song\inst{1} \and
Xiaoyi Zeng\inst{1} \and
Chenqu Ren\inst{2}\and
Juyong Zhang\inst{1}}

\authorrunning{F.~Author et al.}

\institute{ University of Science and Technology of China \and East China Normal University \\ \vspace{2em} \url{https://ustc3dv.github.io/City-on-Web/}}

\maketitle
\vspace{-1em}

\begin{abstract}

Existing neural radiance field-based methods can achieve real-time rendering of small scenes on the web platform. However, extending these methods to large-scale scenes still poses significant challenges due to limited resources in computation, memory, and bandwidth. In this paper, we propose City-on-Web, the first method for real-time rendering of large-scale scenes on the web. We propose a block-based volume rendering method to guarantee 3D consistency and correct occlusion between blocks,  and introduce a Level-of-Detail strategy combined with dynamic loading/unloading of resources to significantly reduce memory demands. Our system achieves real-time rendering of large-scale scenes at approximately 32FPS with RTX 3060 GPU on the web and maintains rendering quality comparable to the current state-of-the-art novel view synthesis methods.

\end{abstract}

\section{Introduction}

\label{sec:intro}

Neural radiance field (NeRF) has significantly advanced the field of scene reconstruction, showing an unparalleled ability to capture complex details across diverse environments. Existing works have demonstrated its ability to render small scenes with exceptional quality and performance in real-time~\cite{cao2023real,yariv2023bakedsdf,yu2021plenoctrees,tang2023delicate,nsvf,reiser2021kilonerf,chen2023mobilenerf,wan2023lduplex,garbin2021fastnerf,bungeenerf,zhang2020nerf++,nerf-w}. NeRF has also been successfully applied to the rendering of large scenes in offline settings, achieving exceptional visual fidelity and generating intricately detailed results~\cite{blocknerf,guo2023streetsurf,bungeenerf,gridnerf,meganerf}.

Despite these successes, real-time neural rendering of large scenes on the web remains profoundly challenging due to inherent computational power, memory, and bandwidth limitations on commodity devices. MERF~\cite{reiser2023merf} has recently achieved significant progress by employing a baking technique to reduce query network calls in the rendering pipeline, thereby enabling real-time rendering of small-scale scenes on the web. However, MERF struggles to capture intricate details in large scenes due to its limited resolution. A naive solution would be to simply increase the volumetric representation's resolution, but this approach would lead to unacceptable increases in memory usage, scaling with $O(N^3)$, and a significant decrease in rendering speed.

To overcome these limitations, we integrate MERF with a block-based strategy~\cite{meganerf} for reconstructing large scenes, a method supported by numerous studies~\cite{blocknerf,meganerf,zhenxing2022switch}. This approach not only improves the model's representational ability but also controls memory growth at an $O(N^2)$ rate because we divide the scene based on ground coordinates without dividing the height. However, there are certain challenges associated with a resource-independent block-based rendering approach on the web.  Specifically, rendering on the web faces limitations on the number and resolution of texture units that can be loaded into a shader (typically no more than 16 texture units), which prevents loading all block resources into a single shader. Consequently, we load the rendering resources of different blocks into their respective shaders. Nevertheless, rendering with different shaders causes issues with 3D consistency. Specifically, when a ray traverses multiple blocks, sampling points might belong to different blocks loaded by different shaders, preventing standard volume rendering. We are thus compelled to render each block sequentially and subsequently combine the rendering results of the different blocks. To this end, we propose a block-based volume rendering strategy and demonstrate that this method of sequential block rendering is equivalent to volume rendering, thereby ensuring correct occlusion and 3D consistency of the rendering results.

Moreover, when viewing from a higher altitude viewpoint, the rendering resources of all scene blocks are needed. Nonetheless, loading all blocks for rendering is impractical due to the excessive memory usage that would surpass the capacity of standard consumer devices. To address this issue, we draw inspiration from traditional graphics techniques~\cite{lodclark1976hierarchical,lodcrassin2009gigavoxels,lodduchaineau1997roaming,lodguthe2002interactive,lodlindstrom2001visualization,lodluebke2003level,lodprogressive} to create Level-of-Detail (LOD) for each block's resources, dynamically selecting resources for rendering based on the camera's position and field of view. This approach significantly reduces the resource demands during rendering, paving the way for smoother user experiences even on less capable devices.

In summary, the contributions of this paper include the following aspects:
\begin{itemize}
    \item[ $\bullet$] We propose a block-based multi-shader volume rendering method, ensuring real-time high-fidelity rendering of large-scale scenes.
    \item[ $\bullet$] We employ an LOD strategy and dynamic loading/unloading strategies to adaptively manage rendering resources and significantly reducing the quantity of resources loaded and ensuring efficient resource utilization for large-scale scene rendering.
    \item[ $\bullet$] Our experiments demonstrate that our system achieves real-time rendering of large-scale scenes at approximately 32FPS with 1080P resolution on an RTX 3060 GPU, while maintaining rendering quality comparable to the current state-of-the-art (SOTA) methods for large-scale scenes.
\end{itemize}

\section{Related Work}
\label{sec:formatting}

\begin{figure*}[t]
    \centering
    \includegraphics[width=1.0 \linewidth]{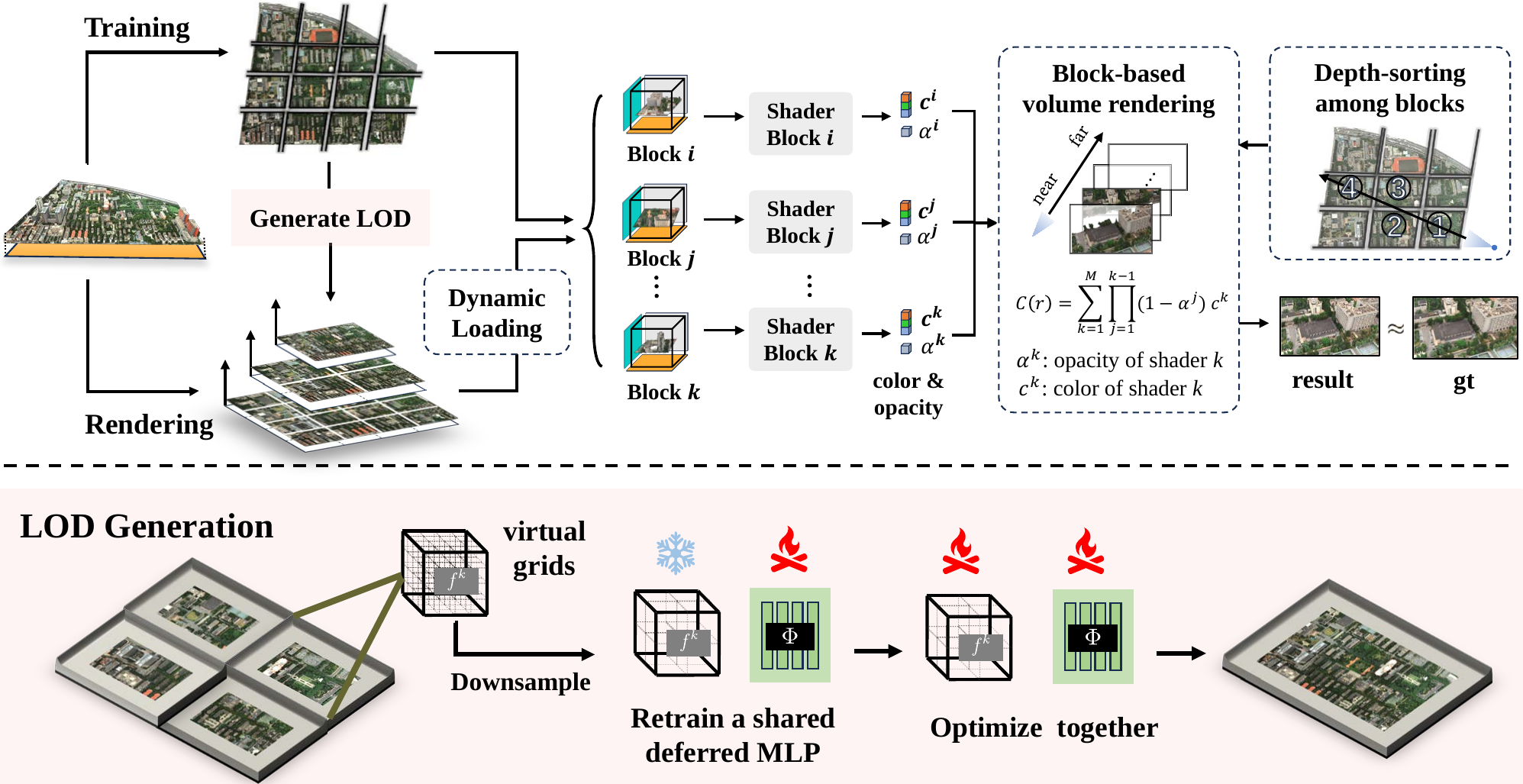}
    \caption{\textbf{Overview of City-on-Web pipeline.} During the training phase, we uniformly partition the scene and reconstruct it at the finest LOD. To ensure 3D consistency, we use a resource-independent block-based volume rendering strategy (\cref{sec:training}). For LOD generation, we downsample virtual grid points and retrain a coarser model (\cref{sec:lod}). This approach supports subsequent real-time rendering by facilitating the dynamic loading of rendering resources.}
  \label{fig:pipeline of training}
  \vspace*{-1.5 em}
\end{figure*}

\textbf{{Large-scale Scene Reconstruction.}} For radiance field reconstruction of large-scale scenes, a key issue lies in enhancing the model’s ability to adequately capture and render extensive scenes. Some works~\cite{blocknerf,meganerf,gu2024ue4} address this by adopting a divide-and-conquer strategy, segmenting expansive scenes into smaller blocks, and applying localized NeRF processing to each. This approach significantly improves both the reconstruction quality and the model's scalability to larger scenes. Switch-NeRF~\cite{zhenxing2022switch} employs a gating network to dispatch 3D points to different NeRF sub-networks. Grid-NeRF~\cite{gridnerf} utilizes a compact multiresolution feature plane and combines the strengths of smoothness from vanilla NeRF with the local detail capturing ability of feature grid-based methods~\cite{muller2022ngp,chen2022tensorf,sun2022direct}, efficiently reconstructing large scenes with fine details. NeRF++~\cite{zhang2020nerf++} enhances the reconstruction of unbounded scenes through its innovative multi-spherical representation. On the other hand, Mip-NeRF 360~\cite{mipnerf360} introduces a scene contraction function to effectively represent scenes that extend to infinity, addressing the challenge of vast spatial extents. F2-NeRF~\cite{wang2023f2} takes this a step further by implementing a warping function for local spaces, ensuring a balance of computational resources and training data across different parts of the scene.

\textbf{{Real-time Rendering.} }Early works mainly focus on the real-time rendering of a simple single object. NSVF~\cite{nsvf} improves NeRF by introducing a more efficient sparse voxel field, significantly accelerating rendering speed while maintaining high-quality output. KiloNeRF~\cite{reiser2021kilonerf} utilizes thousands of small MLPs, each responsible for a tiny scene region, significantly reducing network evaluation time. In contrast, SNeRG~\cite{hedman2021baking} leverages pre-computed sparse grids, allowing for direct retrieval of radiance field information without needing network evaluation. Termi-NeRF~\cite{piala2021terminerf} terminates ray marching in less impactful scene regions, slashing computation time. DONeRF~\cite{neff2021donerf} focuses on one sample using a depth oracle network, speeding up rendering while preserving scene quality. Recently, there have been developments that enable real-time rendering of neural radiance fields in small scenes. MERF~\cite{reiser2023merf} improves upon SNeRG by utilizing a voxel and triplane hybrid representation to reduce memory usage. MobileNeRF~\cite{chen2023mobilenerf} introduces the polygon rasterization rendering  pipeline, running NeRF-based novel view synthesis in real-time on mobile devices. BakedSDF~\cite{yariv2023bakedsdf} bakes volumetric representation into meshes and utilizes spherical harmonics for representing view-dependent color, while NeRF2Mesh~\cite{tang2023delicate} iteratively refine both the geometry and appearance of the mesh. Furthermore, several methods~\cite{turki2024hybridnerf,adaptiveshells2023} exploit the real-time rendering attributes of mesh representations alongside the robust representational potential of volume representations, particularly for rendering hair, translucent materials, and similar entities. These hybrid methods facilitate the achievement of high-fidelity real-time rendering. Recently, 3D Gaussian splatting~\cite{kerbl3Dgaussians} achieves real-time rendering by utilizing a novel 3D Gaussian scene representation and a rasterization-based rendering pipeline. However, extending this representation to large scenes is challenging due to its substantial memory consumption.

\textbf{{Level of Detail.} }Substantial works are devoted to integrating LOD methods into the fabric of traditional computer graphics~\cite{lodclark1976hierarchical,lodcrassin2009gigavoxels,lodduchaineau1997roaming,lodguthe2002interactive,lodlindstrom2001visualization,lodluebke2003level,lodprogressive,losasso2004geometry}, aiming to streamline rendering processes, reduce memory footprint, bolster interactive responsiveness.  Recently, some works have begun to apply LOD to neural implicit reconstruction.  NGLoD~\cite{takikawa2021lod} represents LOD through a sparse voxel octree, where each level of the octree corresponds to a different LOD, allowing for a finer discretization of the surface and more detailed reconstruction as the tree depth increases. Takikawa \etal ~\cite{takikawa2022variable} efficiently encode 3D signals into a compact, hierarchical representation using vector-quantized auto decoder method. BungeeNeRF~\cite{bungeenerf} employs a hierarchical network structure, where the base network focuses on learning a coarse representation of the scene, and subsequent residual blocks are tasked with progressively refining this representation. TrimipRF~\cite{hu2023tri} and LOD-Neus~\cite{zhuang2023anti} leverage multi-scale triplane and voxel representations to capture scene details at different scales, effectively implementing anti-aliasing to enhance the rendering and reconstruction quality.

\section{Background}

Our exploration begins with an in-depth analysis of two influential works, SNeRG~\cite{hedman2021baking} and MERF~\cite{reiser2023merf}, which have both set benchmarks for real-time rendering of the radiance field. SNeRG precomputes and stores a Neural Radiance Fields model in a sparse 3D voxel grid. Each active voxel in SNeRG contains several attributes: density, diffuse color, and  specular feature vector that captures view-dependent effects. Additionally, an indirection grid is used to enhance rendering by either indicating empty macroblocks or pointing to detailed texels in a 3D texture atlas. This representation allows real-time rendering on standard laptop GPUs. 

The indirection grid assists in raymarching through the sparse 3D grid by passing empty regions and selectively accessing non-zero densities $\sigma_i$, diffuse colors $\boldsymbol{c}_i$, and feature vectors $\boldsymbol{f}_i$ from baked textures. Integrating along each ray $\boldsymbol{r}(t) = \boldsymbol{o} + t\boldsymbol{d}$, they compute the sum of the weights, which can be considered as the pixel's opacity:

\begin{equation}
{{\alpha}}(\boldsymbol{r}) = \sum_i w_i, \  w_i=\prod_{j=1}^{i-1}(1-\alpha_j)\alpha_i, \alpha_i=1-e^{-\sigma_i\delta_i}.
\label{eq:alpha_accumulation}
\end{equation}
The step size $\delta_i$ during ray marching  is equal to the voxel width for an occupied voxel. The color $\boldsymbol{C}_d(\boldsymbol{r})$ and specular feature $\boldsymbol{F}_s(\boldsymbol{r})$ along the ray are accumulated using the same weights to compute the final diffuse color and specular feature of ray:

\begin{equation}
{\boldsymbol{C}}_d(\boldsymbol{r}) = \sum_i w_i \boldsymbol{c}_i, \quad {\boldsymbol{F}}_s(\boldsymbol{r}) = \sum_i w_i \boldsymbol{f}_i.
\label{eq:combined_accumulation}
\end{equation}

 Subsequently, the accumulated diffuse color and specular feature vector, along with the positional encoding $PE(\cdot)$ of the ray's view direction, are concatenated to pass through a lightweight deferred MLP $\Phi$ to produce a view-dependent residual color:
\begin{equation}
{\boldsymbol{C}}(\boldsymbol{r})=\boldsymbol{C}_d + \Phi(\boldsymbol{C}_d , \boldsymbol{F}_s, PE(\boldsymbol{d})).
  \label{eq: deferred}
\end{equation}

While SNeRG achieves impressive real-time rendering results, its voxel representation demands substantial memory, which poses limitations for further applications. MERF presents a significant reduction in memory requirements in comparison to extant radiance field methods like SNeRG. By leveraging hybrid low-resolution sparse voxel and 2D high-resolution triplanes, MERF optimizes the balance between performance and memory efficiency. Moreover, it incorporates two pivotal strategies to bridge the gap between training and rendering performance. MERF simulates finite grid approach during training, querying MLPs at virtual grid corners and simulates quantization during training to mimic  the rendering pipeline closely.



\section{Method}
\label{sec:method}

In this section, we present a method for representing and rendering large scenes on the web. Our approach utilizes a block and LOD strategy for rendering large-scale scenes, as described in (\cref{sec:representation}). A block-based volume rendering approach is introduced for the seamless blending of different blocks, utilized both during the training and rendering stages to ensure consistency (\cref{sec:training}).~\cref{sec:optim} details the optimization strategies. The generation and refinement of LODs~\cref{sec:lod} are also explained. ~\cref{sec:baking} elucidates the baking strategy suitable for the representation.
\subsection{Large-scale Radiance Field} 
\label{sec:representation}

Although real-time rendering methods like MERF can achieve high-quality real-time rendering for small-scale scenes, they face representational capacity challenges when applied to larger scenes. As mentioned in \cref{sec:intro}, utilizing a single MERF model to represent vast scenes is problematic due to its limited resolution, especially in terms of detailed and accurate reconstruction. Therefore, we represent scenes using multiple blocks. However, this approach necessitates employing an LOD strategy to reduce the number of resources that need to be loaded during the rendering phase. Thus, we adopt a block-based and LOD strategy for representing the whole scene in the rendering stage. We will elaborate on the  representation used in the rendering stage and provide the representation used to reconstruct the scene in the finest LOD in the training stage.

\textbf{Training Stage.} In the training stage, we only represent and optimize the finest LOD. We uniformly partition the entire scene into $K$ blocks $\{\mathcal{B}_k\}_{k=1}^{K}$, each centered at $\mathbf{c}_k = (x_k, y_k)$, on the $xy$ plane (i.e., the ground plane). This approach stems from the observation that large scenes typically exhibit smaller scales in the $z$-direction compared to the $xy$-plane, prompting us to partition based on the ground plane and avoid subdividing along the $z$-axis. For a point \(\mathbf{p} = (p_x, p_y, p_z) \in \mathbb{R}^3\), we determine its corresponding block $\mathcal{B}_k$ based on its \(xy\) coordinates, denoted as \(\mathbf{p}_{proj} = (p_x, p_y)\):
\begin{equation}
\mathbf{p} \in \mathcal{B}_k, k = \mathop{\arg\min}\limits_{k}\| \mathbf{p}_{proj} - \mathbf{c}_k \|_{\infty}
\label{eq:blocks}
\end{equation}
 Within block $k$, the following trainable components are introduced: (1) $f^{k}$ is an attribute query function that adopts a hash encoding and an MLP decoder that outputs attributes of points such as densities, diffuse color and specular feature (2) $\Phi^{k}$ is a tiny deferred MLP account for view-dependent effects. (3) $\psi^{k}$ is a proposal MLP for sampling.
 

\textbf{Rendering Stage.} In the rendering stage, our scene representation includes hierarchical $L$ LODs representation for the scene. Specifically, as shown in the right figure, we merge $2\times 2$ blocks into one block between two consecutive LODs. As a result, for LOD  \(l\), where \(l \in \{1, 2, \ldots, L\}\), there are \(K/4^{l-1}\) blocks. In each block, the following baked textures are used for rendering: (1) \(f^{k,l}\) is an attribute query function that takes the coordinates of a sample point as input and directly accesses the opacity, diffuse color  and specular features of the sample point from the baked sparse voxel and triplane textures. (2) \(\Phi^{k,l}\) represents a tiny deferred MLP that accounts for view-dependent effects. (3) \(\psi^{k,l}\) is used as a multi-level occupancy grid for sampling.

 \subsection{Block-based Volume Rendering} 
 
 \label{sec:training}

\begin{figure}[t]
    \centering
    \begin{subfigure}[t]{0.35\columnwidth}
        \centering
        \includegraphics[height=4.5cm]{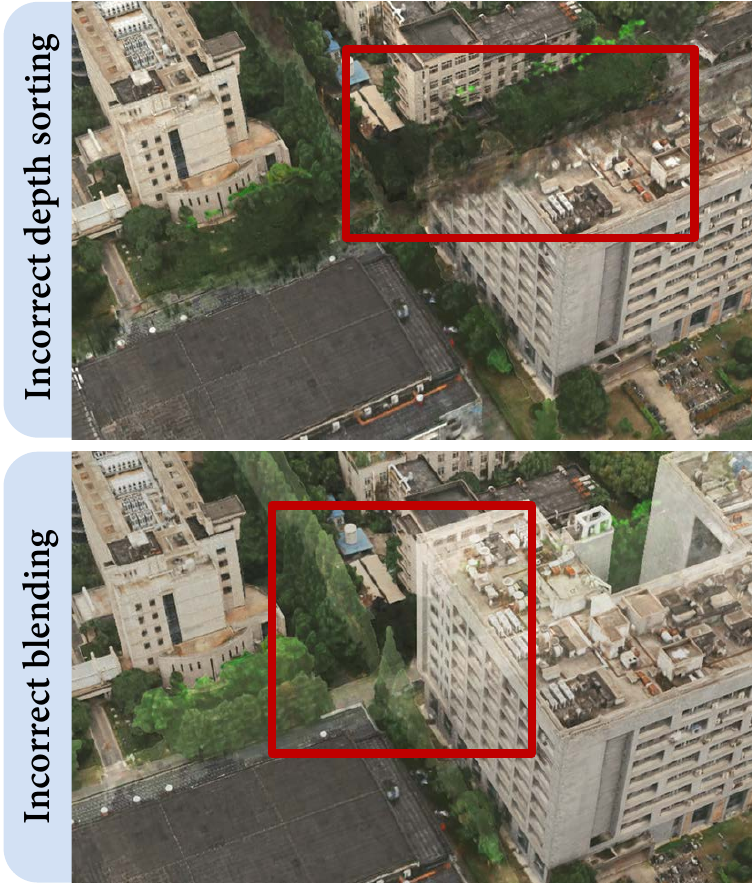}
        \caption{Incorrect blending}
        \label{fig:problem}
    \end{subfigure}
    \hspace{0.0em} 
    \rotatebox{90}{\hdashrule{4.6cm}{0.6pt}{1mm 1mm}}
    \hspace{0.3em} 
    \begin{subfigure}[t]{0.55\columnwidth}
        \centering
        \includegraphics[height=4.5cm]{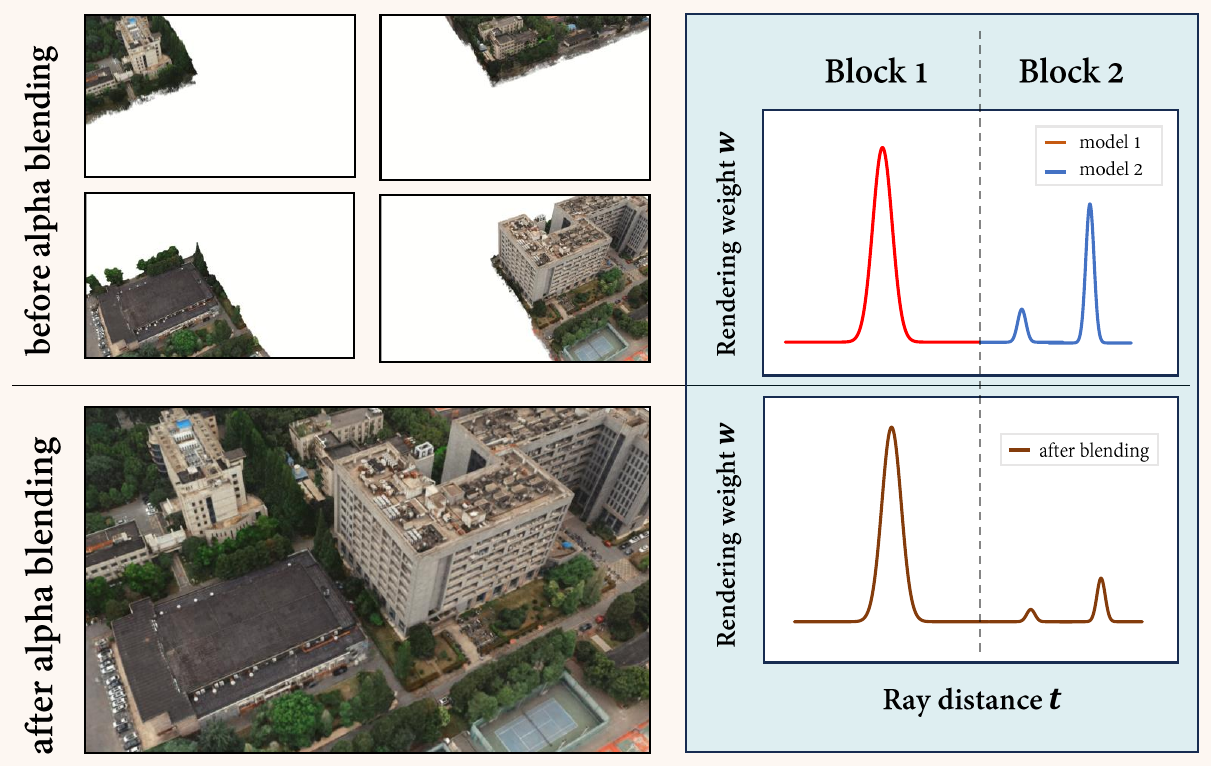}
        \caption{Ours alpha blending}
        \label{fig:consistent training}
    \end{subfigure}
        \caption{\textbf{Visualization comparison between the alpha blending method and others.} (a) Top image: incorrect occlusion without depth sorting. Bottom image:  incorrect rendering results when simply using  $\alpha_i/(\sum_j \alpha_{j})$ as blending weights. (b) Left: rendering results of four separate blocks and the final blending result. Right: visualization of sample points' rendering weights before and after alpha blending.} 
  \label{fig:alpha blending}
  \vspace*{-1em}
\end{figure}

In the rendering stage, we create multiple shaders to render distinct blocks. Specifically, one shader is allocated for storing the texture of an individual block. Each block subsequently renders an image respective to the current camera view. However, a simplistic averaging of these resultant rendering outputs can lead to discernible seams and does not ensure correct occlusion at the inter-block boundaries as shown in~\cref{fig:problem}. Therefore, we employ a block-based volume rendering strategy and combine it with depth sorting followed by alpha blending to ensure seamless boundaries and correct occlusion at the edges. 

Specifically, in the rendering stage, for a ray $\boldsymbol{r}(t)$ passing through $\boldsymbol{M}$ blocks with a total of $\boldsymbol{N}$ samples, where each block $k$ has $\boldsymbol{n}_k$ samples, we perform volume rendering within each block to obtain its individual rendering diffuse color $\boldsymbol{C}_d^{k}$, specular feature $\boldsymbol{F}^{k}$ and opacity ${\alpha}^{k}$ of the ray in block $\mathcal{B}_k$  according to ~\cref{eq:alpha_accumulation,eq:combined_accumulation}. Then we get  final rendering color $\boldsymbol{C}^{k}$ of block $\mathcal{B}_k$ according to~\cref{eq: deferred}. Subsequently, to correctly handle occlusion in rendering, we depth-sort the blocks and apply volume rendering across multiple blocks in sequence, using opacity to generate the blending weights:
\begin{equation}
\begin{aligned}
    \boldsymbol{C}(\boldsymbol{r}) = \sum_{k=1}^M \prod^{k-1}_{j=1}(1-\alpha^{j})\boldsymbol{C}^k.
\end{aligned}
\label{eq:volume rendering across blocks}
\end{equation}

\begin{wrapfigure}{r}[1em]{0.7\columnwidth}
\vspace*{-2.2em}
    \includegraphics[width=0.95\linewidth]{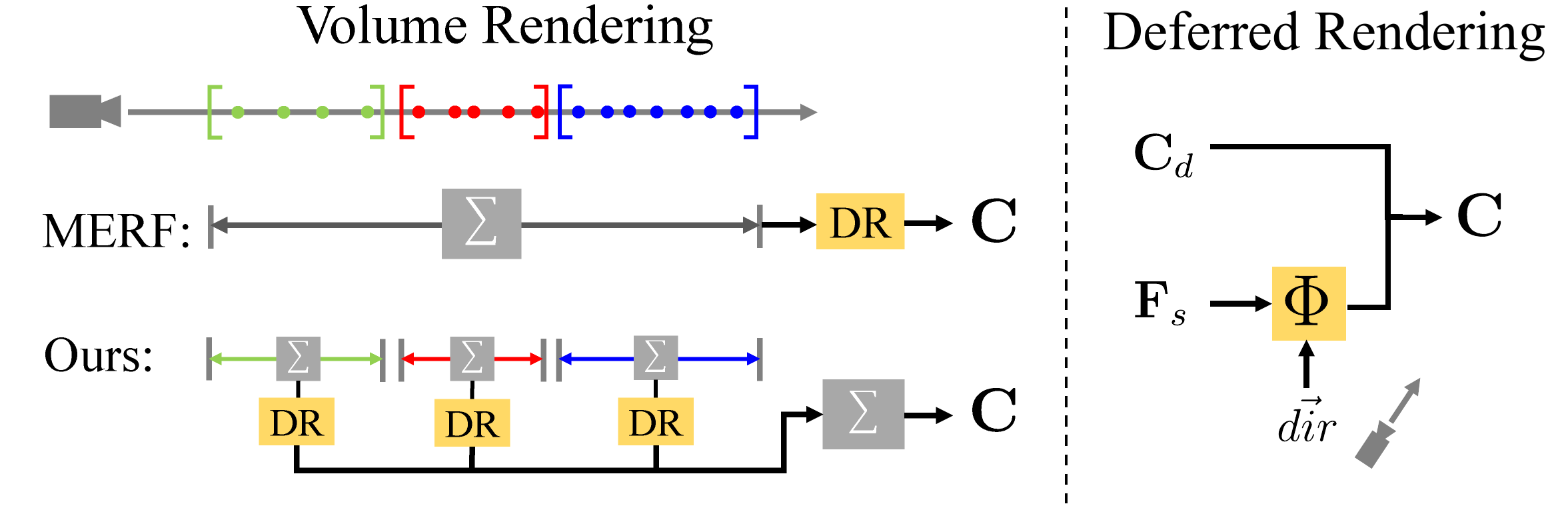}
    \caption{\textbf{Block-based volume rendering.}  "DR" denotes deferred rendering. $\Phi$ represents the deferred MLP.}
  \label{fig:vr}
  \vspace*{-2.5em}
\end{wrapfigure}
Under the Lambertian surface setting where the specular color is zero, the  color obtained from volume rendering on the total of $\boldsymbol{N}$ ray samples from~\cref{eq:combined_accumulation} is equal to the results produced by our approach of conducting volume rendering within each block followed by inter-block volume rendering~\cref{eq:volume rendering across blocks}. The proof is given in the supplementary. Thus, our rendering approach maintains correct occlusion and keeps 3D consistency when using multiple shaders rendering on the web as shown in~\cref{fig:consistent training}.


The volume rendering process in MERF involves integrating all sample points together, followed by deferred rendering. In contrast, as shown in~\cref{fig:vr}, our block-based volume rendering is fundamentally based on segmented integration. Without the deferred rendering process, it is entirely equivalent to traditional volume rendering. However, if we adhere to the MERF rendering pipeline during the training process, it will lead to discrepancies between the rendering results during the training and rendering phases, ultimately affecting the rendering quality. To minimize this gap, we adopt the same rendering pipeline during the training stage as we do in the rendering stage. 

Specifically, in the training stage, for ray \(\boldsymbol{r}(t)\), we uniformly sample between the near and far boundaries based on the scene's bounding box. Then, we determine that this point is inside $\mathcal{B}_k$ according to~\cref{eq:blocks} and query the corresponding proposal MLP $\psi^{k}$ of $\mathcal{B}_k$ to sample probability distributions along the ray. Similarly, we also query the corresponding \(f^{k}\) to obtain the attributes of the rendering sample points. Lastly, like in the rendering stage, we render each block sequentially to obtain the color and opacity for the ray in $\mathcal{B}_k$ and use ~\cref{eq:volume rendering across blocks} to derive the final rendering result.

\subsection{Optimization}
\label{sec:optim}

In the training stage, we reconstruct the finest LOD model by optimizing it with various losses:
\begin{equation}
\mathcal{L}_{\text{train}} = \mathcal{L}_{\text{cb}} + \mathcal{L}_{\text{global}} + \lambda_1 \mathcal{L}_{\text{s3im}} + \lambda_2 \mathcal{L}_{\text{prop}} + \lambda_3 \mathcal{L}_{\text{dist}} + \lambda_4 \mathcal{L}_{\text{s}} + \lambda_5 \mathcal{L}_{\text{opacity}}.
\label{eq:total_loss}
\end{equation}

Here, we use Charbonnier loss~\cite{charbonnier1997deterministic} $\mathcal{L}_{cb}$ for reconstruction and S3IM loss~\cite{xie2023s3im} $\mathcal{L}_{s3im}$ to assist model in capturing high-frequency details. Additionally, we use the interlevel loss $\mathcal{L}_{prop}$ to provide a supervision signal for proposal MLP and distortion loss $\mathcal{L}_{dist}$ to reduce floaters like Mip-nerf 360~\cite{mipnerf360}.

\textbf{Sparsity Loss.} We random uniform sample points set $\mathcal{P}$ within the bounding box of the scene and apply $L_{1}$ regularization on the opacity of sample points $\alpha_i$ to encourage model to predict sparse occupied space:
\begin{equation}
{\mathcal{L}_{\textrm{sparse}}} = \frac{1}{|\mathcal{P}|}\sum_{p_i\in \mathcal{P}}|\alpha_i|
  \label{eq:sparse loss}
\end{equation}

\textbf{Opacity Loss.} We introduce a regularization term for the opacity of the block. This regularization encourages the opacity of the block to be as close to 0 or 1 as possible, implying either full transparency or full opaqueness:
\begin{equation}
{\mathcal{L}_{\textrm{opacity}}}= -\sum_{k}(\alpha^{k}log(\alpha^{k}) + (1-\alpha^{k})log(1-\alpha^{k})).
  \label{eq:opacity loss}
\end{equation}

\textbf{Regularization of Deferred MLPs.}
 In the deferred rendering context, various combinations of specular and diffuse colors can satisfy multi-view consistency constraints. This situation often leads to incorrect disentanglement of these color components. Our training process involves using a deferred MLP within each block, but this approach does not guarantee the smoothness of specular color across block boundaries and the multitude of possible combinations of specular and diffuse colors. 
 
 Inspired by Grid-NeRF~\cite{gridnerf}, which utilizes the smoothness of MLP to regularize explicit grid representations. We also utilize a global deferred MLP to regularize the rendering outputs from smaller, block-specific deferred MLPs, ensuring the global smoothness of specular color. In particular, we combine the specular color generated by this global deferred MLP with the diffuse color to obtain the final rendering result. We then supervise the rendering result using ground truth images in the form of a Charbonnier loss, denoted as $\mathcal{L}_{global}$, to regularize the smaller deferred MLPs. Notably, this global deferred MLP is significantly larger and thus possesses sufficient representational capacity compared to the smaller deferred MLPs designated for each block. Therefore, the global MLP does not limit the model's representational capacity.


\subsection{LOD Generation}
 \label{sec:lod}
To ensure high-quality rendering from elevated viewpoints and reduce resource usage for distant scene blocks, our method generates multiple LODs for the scene. One conventional approach to generate LODs would be to retrain the entire scene using fewer blocks, that is, at a lower representation resolution, but this method  extends the training time a lot. Additionally, considering the specialized photography techniques employed for capturing large scenes, usually from aerial or top-down perspectives, it is challenging to ensure appearance consistency across models trained separately for extrapolated views if we retrain the entire scene from scratch.

Therefore, we generate LODs based on the scene's finest LOD acquired during the training stage. Specifically, we simulate the virtual grid  to store rendering attributes like MERF in the training stage. As merging \(M \times M\) blocks into $\frac{M}{2}  \times \frac{M}{2}$ blocks to generate  LODs, we initially downsample the resolution of the virtual grid in each block by a factor of 2. Subsequently, we freeze the training of the query function \(f^k\) within these submodels and retrain a new shared deferred MLP \(\Phi^{k,l}\) across merged blocks. Finally, we continue to jointly optimize these submodels and the deferred MLP to adapt to lower-resolution voxels and triplanes.

\subsection{Baking}
 \label{sec:baking}
 For LOD $l$,  we merge \(M \times M\) blocks into  $\frac{M}{2^l} \times \frac{M}{2^l}$ blocks. Every \(2^l \times 2^l\) blocks can be baked into a single texture asset $f^{k,l}$ in the baking stage thanks to the dowmsampling when generating LODs. Thus, in the rendering stage, a single shader is responsible for rendering these \(2^{l} \times 2^{l}\) blocks.

Specifically, we render all training rays to collect ray samples initially. Samples with opacity and weight values above a certain threshold are retained, and samples below the threshold are discarded. The preserved samples are used to mark the adjacent eight grid points as occupied in the binary occupancy grids $\phi^{k,l}$. After generating binary grids to identify occupied voxels, we follow  MERF by baking high-resolution 2D planes and a low-resolution 3D voxel grid in each block to get the attribute function $f^{k,l}$ used in the rendering stage. Only the non-empty 3D voxels are stored using a  block-sparse format. We downsample the occupancy grid with max-pooling for efficient rendering. To further save storage, we compress textures into the PNG format.

\section{Experiments}

\begin{table*}[t]
\centering
\footnotesize
\renewcommand\arraystretch{1.5}
\caption{\textbf{Quantitative comparison on the \textit{Matrix City}, \textit{Campus}, and \textit{Rubble} datasets.} We report PSNR, LPIPS, and SSIM on the test views. The \textbf{best} and \underline{second best} results are highlighted.}
\begin{tabular}{@{}l|ccc|ccc|ccc@{}}
\toprule[1.5pt]
 & \multicolumn{3}{c|}{\textit{Matrix City}} & \multicolumn{3}{c|}{\textit{Campus}} & \multicolumn{3}{c}{\textit{Rubble}} \\

 & PSNR$\uparrow$ & LPIPS$\downarrow$ & SSIM$\uparrow$ & PSNR$\uparrow$ & LPIPS$\downarrow$ & SSIM$\uparrow$ & PSNR$\uparrow$ & LPIPS$\downarrow$ & SSIM$\uparrow$ \\
\midrule[1.0pt]
NeRFacto & 24.95 & \underline{0.456} & 0.688 & \underline{23.47} & \underline{0.255} & \underline{0.689} & 19.02 & \underline{0.538} & 0.512 \\
Instant-NGP & 23.55 & 0.597 & 0.629 &  21.91 & 0.478 & 0.549 & 20.37 & 0.629 & 0.478 \\
Mega-NeRF & \underline{25.43} & 0.517 & 0.674 & 22.28 & 0.472 & 0.565 & \textbf{23.68} & 0.558 & \underline{0.525} \\
Grid-NeRF & {24.90} & 0.480 & \underline{0.698} & - & - & - & - & - & - \\
Ours & \textbf{25.87} & \textbf{0.332} & \textbf{0.734} & \textbf{24.73} & \textbf{0.192} & \textbf{0.736} & \underline{21.32} & \textbf{0.482} & \textbf{0.539} \\
\bottomrule[1.5pt]
\end{tabular}
\label{tab:regix}
\vspace*{-0.5em}
\end{table*}


\subsection{Experiments setup}

\begin{figure*}
    \centering
    \includegraphics[width=1.0\linewidth]{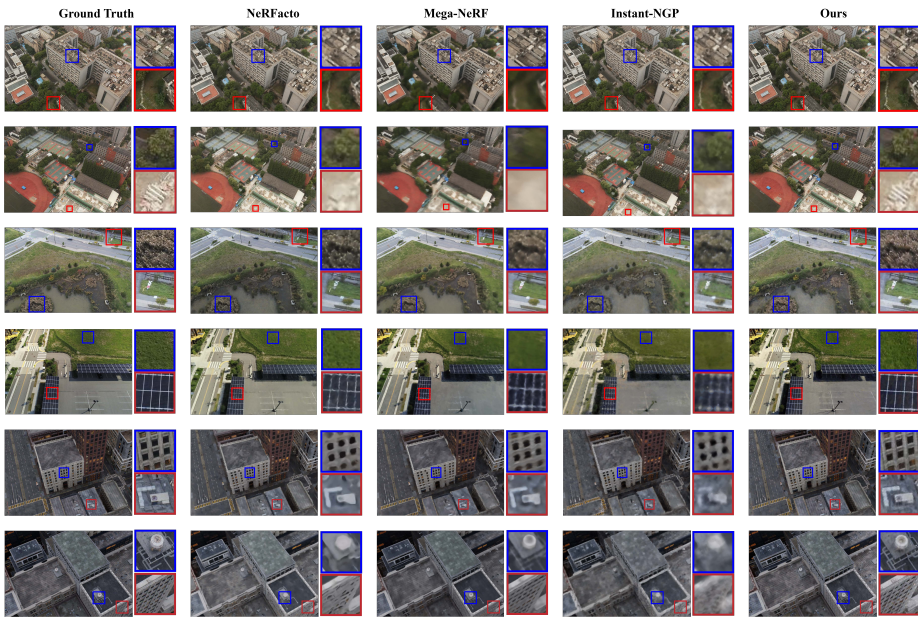}
    \caption{\textbf{Qualitative comparisons with existing SOTA methods.} By testing different methods across diverse scales and environments, it clearly reveals that our approach excels in recovering finer details and achieves a higher quality of reconstruction.}
    \label{fig:sota}
\end{figure*}


\textbf{Dataset and Metric.} Our experiments span across various scales and environments. We have incorporated a real-world urban scene dataset (\textit{Campus}) and public datasets consisting of real-world rural rubble scenes (\textit{Rublle})~\cite{meganerf} and synthetic city-scale data (\textit{MatrixCity})~\cite{li2023matrixcity}. Our datasets were recorded under uniform, cloudy lighting conditions to minimize variation. To obtain precise pose information, we employed an annular capturing approach, which has a higher overlap rate compared to grid-based capturing methods. The dataset covers an area of $1200\times 800$ square meters. It includes a total of 6515 images. We use $99\%$ data for training, and the rest is used as the test dataset. To assess the quality and fidelity of our reconstructions, we employ various evaluation metrics, including \textbf{PSNR}, \textbf{SSIM }and \textbf{LPIPS}~\cite{zhang2018unreasonable}.

\noindent \textbf{Implementations and Baselines.} Our method takes posed multi-view images captured using a fly-through camera as input. The training code is built on the nerfstudio framework~\cite{tancik2023nerfstudio} with tiny-cuda-nn~\cite{tiny-cuda-nn} extension. Our real-time viewer is a JavaScript web application whose rendering is implemented through GLSL. We set the $512^3$ resolution for the voxel and $2048^2$ resolution for the triplane within each block. We use a 4-layer MLP with 64 hidden dimensions as an encoder after multi-resolution hash encoding to output density, color, and specular feature. Moreover, a 3-layer MLP with 16 hidden dimensions tiny deferred MLP is developed to predict residual view-dependent color. We sample $16384$ rays per batch and use Adam optimizer with an initial learning rate of $1\times10^{-2}$ decaying exponentially to $1\times10^{-3}$. The global deferred MLP is a 6-layer MLP with 128 hidden dimensions. Our model is trained with $50k$ iterations on one NVIDIA A100 GPU. We split the scene into 24  blocks for \textit{Campus} scene, and split other scenes into four blocks to reconstruct these scenes. We perform qualitative comparisons between our method and existing SOTA methods for large-scale reconstruction. The \textit{Campus} dataset is partitioned into six parts. NeRFacto, Instant-NGP, and Grid-NeRF were applied to reconstruct one of these parts. NeRFacto and Instant-NGP are utilized with the highest hash encoding resolution of $8192^3$. Mega-NeRF divides the \textit{Campus} scene into 24 blocks and splits another dataset into four blocks to evaluate its performance. 

Our experiments focus on a single part in \textit{Campus} dataset for comparative analysis with existing real-time rendering methods. 
This part contains over 1600 images, covering an area of approximately 
$600\times400$ square meters. We divide this part of the data into four parts for a fair comparison with the MERF method. Other methods did not divide the dataset when conducting experiments. Moreover, we benchmark current real-time rendering methods using three critical parameters: Peak GPU memory usage ({VRAM}), frames per second ({FPS}), and on-disk storage ({DISK}). We report these metrics based on tests conducted on an NVIDIA RTX 3060 GPU with $1920 \times 1080$ resolution.

\subsection{Results Analysis}

\begin{wrapfigure}{r}{0.6\columnwidth}
    \vspace*{-3em}
    \centering
    \includegraphics[width=0.95 \linewidth]{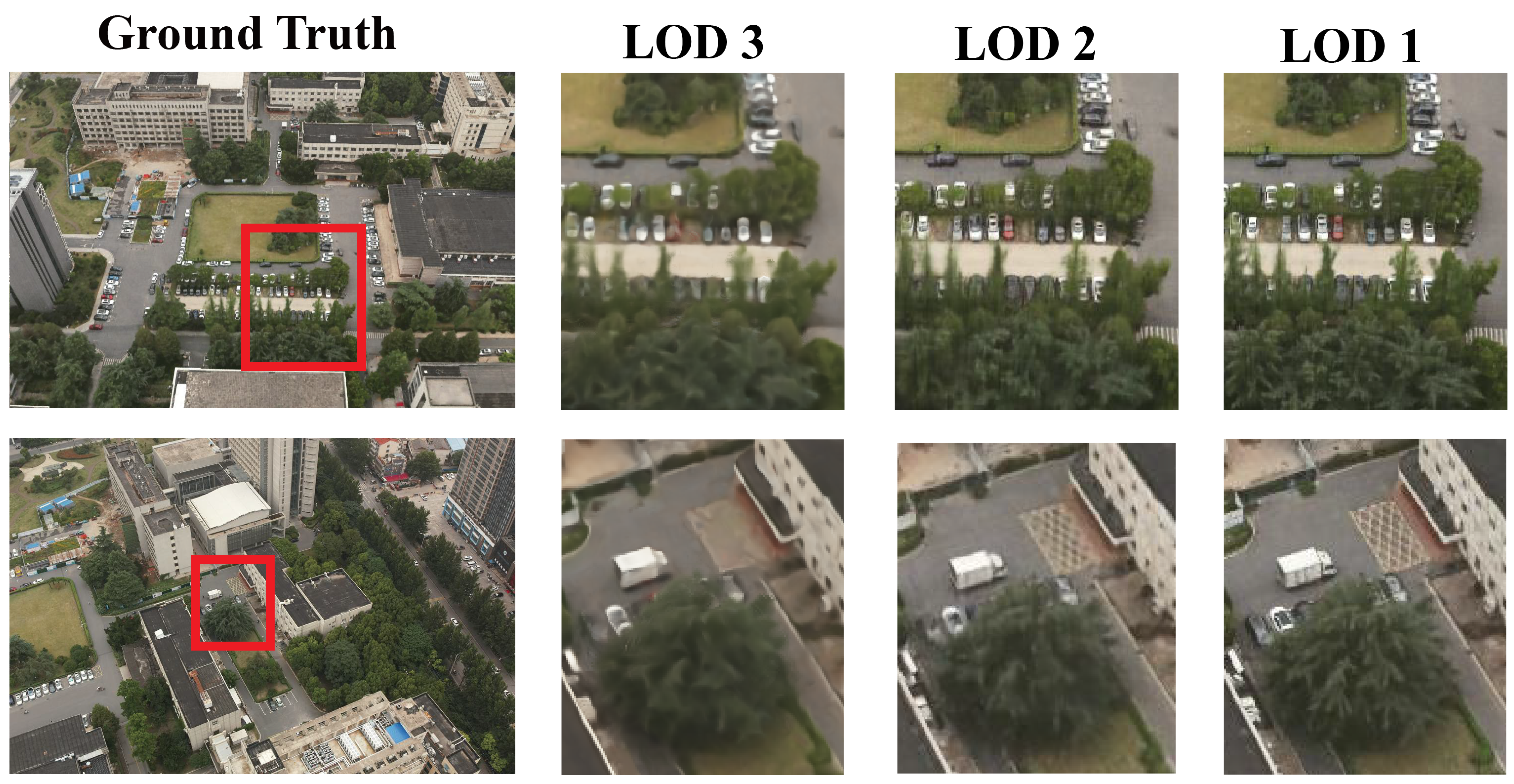}
    \caption{\textbf{Visualization of our LOD result.}}
  \label{fig:lod}
  \vspace*{-1.5em}
\end{wrapfigure}

We systematically evaluate the performance of both baseline models and our method through qualitative and quantitative comparisons in \cref{tab:regix} and \cref{fig:sota}. Notably, our method demonstrates a remarkable enhancement in visual fidelity as reflected by the SSIM and LPIPS metrics, which indicate the extent of detail restoration. Despite a reduction in PSNR compared to the SOTA methods, this is attributable to the fact that LPIPS and SSIM are more sensitive to the recovery of fine details, whereas PSNR mainly measures pixel-wise color accuracy. Our approach achieves higher fidelity reconstructions, revealing finer details due to our partitioned reconstruction strategy. 


\begin{table}
    \centering
 
    \renewcommand{\arraystretch}{1.2}
        \caption{\textbf{Comparison with existing real-time methods.} We divide the scene into four blocks and 16 blocks. 
We split the data into four parts, with each part being reconstructed by an individual MERF for a fair comparison. We present the results of Gauss splatting~\cite{kerbl3Dgaussians}  after training 30,000 and 100,000 iterations for demonstration.  VRAM and DISK are denoted in megabytes (MB).}
    \begin{tabular}{>{\raggedright\arraybackslash}p{3.0cm}|>{\centering\arraybackslash}p{1.2cm}>{\centering\arraybackslash}p{1.2cm}>{\centering\arraybackslash}p{1.2cm}>{\centering\arraybackslash}p{1.2cm}>{\centering\arraybackslash}p{1.2cm}>{\centering\arraybackslash}p{1.2cm}}
    \toprule[1.0pt]
        & PSNR$\uparrow$  & SSIM$\uparrow$ & LPIPS$\downarrow$ & VRAM$\downarrow$ & DISK$\downarrow$ & FPS$\uparrow$ \\
        \midrule[0.7pt]
       MobileNeRF & 19.99 & 0.516 & 0.712 & 712.3 & 242.1 & 68 \\
       BakedSDF & 22.24 & 0.627 & 0.413 & 544.8 & 515.3 & \textbf{223} \\
       MERF(4 blocks) & 24.02 & 0.713 & 0.254 & 592.4  & 121.2 & 58 \\
       GS(3w iters) & 23.78 &  0.745 &  0.263 & 1469.3 & 1469.1 & 77 \\
       GS(10w iters) & 24.94 & \textbf{0.783 }& 0.227 & 1467.1  & 1467.1 & 58 \\
    \midrule[0.7pt]
       Ours(4 Blocks) & 24.82 & 0.741 & 0.190 & \textbf{526.6} & \textbf{114.4} & 46 \\
       Ours(16 Blocks) &\textbf{ 25.13} &  0.779 & \textbf{0.167}  & 2040.7 & 464.5  & 34 \\
    \bottomrule[1.0pt]
    \end{tabular}

    \label{tab:realtime}
\end{table}

In our evaluation, detailed in~\cref{tab:realtime}, we compare our method with current real-time rendering methods, using one part of the~\textit{Campus} dataset for testing. These tests, are performed on an NVIDIA RTX 3060 Laptop GPU at a $1920\times1080$ resolution. The results demonstrate that our method excels in reconstruction quality. We represent each scene block using voxels and triplanes, and store the baked grid attributes as images. This strategy significantly reduces the memory. This reduction notably accelerates resource transmission for web-based rendering applications. However, it is observed that our frame rate during rendering is lower compared to other methods. This is attributed to their rendering pipeline based on mesh rasterization, which is in contrast to our method, which utilizes volume rendering.

\subsection{LOD Result}

 \cref{tab:lod} presents the quantitative rendering results at various LODs, along with the corresponding DISK and VRAM usage. With increasing LOD, the resources required for rendering significantly decrease. Notably, our method's lowest LOD  still maintains high-fidelity rendering results, as demonstrated in~\cref{fig:lod}. Our LOD strategy significantly streamlines the management of resource loading on web platforms, which is particularly advantageous in rendering distant blocks, as it requires less VRAM. It is worth noting that the VRAM usage presented in~\cref{tab:lod} represents the cumulative memory consumption of all blocks. Our dynamic loading strategy adaptively selects resources to load based on the camera's field of view and the distance to each block, effectively keeping the peak VRAM usage around 1100MB.

\begin{table}
    \centering
    \caption{\textbf{The LOD results on the whole \textit{Campus} dataset.}  VRAM and DISK are denoted in megabytes (MB).}
    \renewcommand{\arraystretch}{1.0} 
    \begin{tabular}{@{}>{\centering\arraybackslash}p{2cm}|>{\centering\arraybackslash}p{1.2cm}>{\centering\arraybackslash}p{1.2cm}>{\centering\arraybackslash}p{1.2cm}>{\centering\arraybackslash}p{1.2cm}>{\centering\arraybackslash}p{1.2cm}@{}}
    \toprule[1.0pt]
        & PSNR$\uparrow$ & SSIM$\uparrow$ & LPIPS$\downarrow$ & VRAM$\downarrow$ & DISK$\downarrow$ \\
        \midrule[0.7pt]
       LOD3 & 23.72 & 0.660 & 0.306 & \textbf{132.1} & \textbf{ 40.2} \\
       LOD2 & 24.23 & 0.682 & 0.297 & 841.6 & 201.7 \\
       LOD1 &\textbf{ 24.73} &\textbf{ 0.736} & \textbf{0.192} & 3970.2 & 1259.6 \\
    \bottomrule[1.0pt]
    \end{tabular}
    \label{tab:lod}
\end{table}

\subsection{Ablation Study}

We conduct ablation studies to demonstrate the impact of the contributions introduced to our method.

 \textbf{Ablation on Our Method.}  In  \cref{tab:Ablation}, we conduct an ablation study of our method on one part of  \textit{Campus} dataset ``ours(4 blocks)'' means we use four blocks with $512^3$ voxel resolution and $2048^3$ triplane resolution for scene reconstruction.  ``ours(16 blocks)'' means we use 16 blocks with $512^3$ voxel resolution and $2048^3$ triplane resolution for scene reconstruction. In ``model with  $1024^3$ Res.'', we train a one block MERF model with $1024^3$ voxel resolution and $4096^2$ triplane resolution.  In ``model with  $2048^3$ Res.'', we train a one block MERF model with $2048^3$ voxel resolution and $8192^2$ triplane resolution. Our method has higher rendering quality and requires less storage space.  In ``no alpha blending'', we instead our alpha blending with simply using  $\alpha_i/(\sum_j \alpha_{j})$ as blending weights. This non-occlusion-aware blending strategy significantly reduces the rendering quality. In  ``no consistent training'', we use MERF's volume rendering pipeline in the training stage. In ``no global deferred MLP'', we remove global deferred MLP. Without the regularization of deferred MLPs, the quality of the reconstruction has decreased. 



\begin{table*}
    \centering
 \vspace*{-1em}
    \renewcommand{\arraystretch}{1.1} 
        \caption{\textbf{Ablation study on our method.} The result is tested on one section of the \textit{Campus} dataset.  VRAM and DISK are denoted in megabytes (MB).}
    \begin{tabular}{@{}>{\raggedright\arraybackslash}p{4cm}|>{\centering\arraybackslash}p{1.2cm}>{\centering\arraybackslash}p{1.2cm}>{\centering\arraybackslash}p{1.2cm}>{\centering\arraybackslash}p{1.2cm}>{\centering\arraybackslash}p{1.2cm}@{}}
    \toprule[1.0pt]
        & PSNR$\uparrow$ & SSIM$\uparrow$ & LPIPS$\downarrow$ & VRAM$\downarrow$ & DISK$\downarrow$ \\
        \midrule[0.7pt]
        model with  $1024^3$ Res.  & 24.05 & 0.710 & 0.201 & 540.9 & 147 \\
        model with  $2048^3$ Res. & 24.42 & 0.751 & 0.184 & 3073.2 & 457 \\
        no alpha blending & 24.03 & 0.684 & 0.345 & 565.4 & 153 \\
        no consistent training & 24.21 & 0.702 & 0.281 & 514.4 & 110 \\
        no global deferred mlp & 24.61 & 0.712 & 0.198 & 536.6 & 126 \\
        ours(4 blocks) & 24.82 & 0.741 & 0.190 & \textbf{526.6} & \textbf{114 }\\
       ours(16 Blocks) &\textbf{ 25.13} & \textbf{ 0.779 }& \textbf{0.167}  & 2040.7 & 464  \\
    \bottomrule[1.0pt]
    \end{tabular}
    \label{tab:Ablation}
\end{table*}

\textbf{Ablation on LOD Generation.} \cref{tab:lod Ablation} shows ablation study on LOD generation. We use our LOD generation method as basline. In ``downsample'', we simply downsample the model without re-optimization. In ``retrain from scratch'', we do not use the finest LOD model to generate LOD. Instead, we trained the same resolution model from scratch. 

\begin{table*}
\vspace*{-1em}
    \centering
    \renewcommand{\arraystretch}{1.1} 
        \caption{\textbf{Ablation study on LOD generation.} }
    \begin{tabular}{@{}>{\raggedright\arraybackslash}p{3cm}|>{\centering\arraybackslash}p{1.2cm}>{\centering\arraybackslash}p{1.2cm}>{\centering\arraybackslash}p{1.2cm}>{\centering\arraybackslash}p{2.5cm}@{}}
    \toprule[1.0pt]
        & PSNR$\uparrow$ & SSIM$\uparrow$ & LPIPS$\downarrow$ & Training Time $\downarrow$ \\
        \midrule[0.7pt]
        downsample & 22.43 & 0.614 & 0.362 &\textbf{ 0hours} \\
        retrain from scratch & 24.17 & 0.722 & 0.224 & 12hours\\
       ours & \textbf{24.20} & \textbf{ 0.724} & \textbf{0.204}  &  4hours  \\
    \bottomrule[1.0pt]
    \end{tabular}
    \label{tab:lod Ablation}
\end{table*}


\section{Conclusion}
\label{sec:conculsion}
In this work, we introduced City-on-Web, which to our knowledge is the first system that enables real-time neural rendering of large-scale scenes on the web using laptop GPUs. Our choice of a block-based volume rendering strategy, tailored for resource-independent web environments, achieves seamless integration between blocks. Our carefully designed LOD generation and refinement strategy support dynamic loading, minimizing necessary resources on the web while ensuring the best visual experience. Extensive experiments have also fully proved the effectiveness of City-on-Web.

%
%
\bibliographystyle{splncs04}
\bibliography{main}

\newpage

\renewcommand{\thesection}{\Alph{section}}

\setcounter{section}{0}
\section{Derivation of Block-based Volume Rendering}

\label{sec:proof}
Traditional volume rendering methods, like those used in Block-NeRF~\cite{blocknerf} and Mega-NeRF~\cite{meganerf}, necessitate accessing radiance and opacity for all sampling points, even if the points belong to different blocks. This requirement means that when the rendering resources of multiple shaders cannot communicate with each other, their volume rendering algorithm cannot work. On the contrary, our block-based volume rendering algorithm is tailored for such resource-independent settings, allowing for the independent rendering of each block before blending their outputs. In this section, we will prove the equivalence between this resource-independent volume rendering  and the traditional volume rendering  under Lambertian conditions.

In traditional volume rendering, consider a ray $\boldsymbol{r}$ with $M$ sampling points. For each sampling point $p_i$, the diffuse color, feature, and opacity are represented as $\boldsymbol{c}_i$, $\boldsymbol{f}_i$, and $\alpha_i$ respectively. The final diffuse color $\boldsymbol{C}_d$ of the ray $\boldsymbol{r}$ can be obtained by 
\begin{equation}
\begin{aligned}
\hat{\boldsymbol{C}}_d(\boldsymbol{r})=\sum_{i=1}^{M}\prod^{i-1}_{j=1}(1-\alpha_j)\alpha_i\boldsymbol{c}_i
\end{aligned}
\label{eq:tradition}
\end{equation}

In block-based volume rendering, let's assume the ray, with $M$ sampling points, traverses through K blocks. Within block $\mathcal{B}_k$, there are $N_k$ sampling points. For sampling point $p_i^k$ in block $\mathcal{B}_k$,  the diffuse color, feature, and opacity output by the shader or submodel of block $\mathcal{B}_k$ are represented as $\boldsymbol{c}^k_i$, $\boldsymbol{f}^k_i$, and $\alpha^k_i$, respectively. 
\begin{equation}
\begin{aligned}
\boldsymbol{c}^k_i = \boldsymbol{c}_{N_1+\cdots+N_{k-1}+i}\\
\boldsymbol{f}^k_i = \boldsymbol{f}_{N_1+\cdots+N_{k-1}+i}\\
\alpha^k_i = \alpha_{N_1+\cdots+N_{k-1}+i}
\end{aligned}
\end{equation}
Then, for this ray, diffuse color $\boldsymbol{C}^k_d$, specular feature $\boldsymbol{F}^k$ and opacity $\alpha^k$ of block $\mathcal{B}_k$ are caculated as follows:
\begin{equation}
\begin{aligned}
\boldsymbol{C}^k_d &=\sum_{i=1}^{N_k} \prod_{j=1}^{i-1}(1-\alpha_j^k)\cdot \alpha_i^k \boldsymbol{c}_i^k \\
\boldsymbol{F}^k &=\sum_{i=1}^{N_k} \prod_{j=1}^{i-1}(1-\alpha_j^k)\cdot \alpha_i^k \boldsymbol{f}_i^k \\
\alpha^k  &=\sum_{i=1}^{N_k} \prod_{j=1}^{i-1}(1-\alpha_j^k)\cdot \alpha_i^k
\end{aligned}
\label{eq:volume rendering inside blocks}
\end{equation}
By blending the rendering results of each block using Eq. (8), the final diffuse color $\boldsymbol{C}_d$ and specular feature $\boldsymbol{F}$ of the ray $\boldsymbol{r}$ can be obtained.

\begin{equation}
\begin{aligned}
\boldsymbol{C}_d(\boldsymbol{r})=\sum_{k=1}^K \prod_{j=1}^{k-1}(1-\alpha^j)\cdot \boldsymbol{C}^k_d \\
\boldsymbol{F}(\boldsymbol{r})=\sum_{k=1}^K \prod_{j=1}^{k-1}(1-\alpha^j)\cdot \boldsymbol{F}^k
\end{aligned}
\label{eq:proof}
\end{equation}
From~\cref{eq:volume rendering inside blocks}, we get the following equation:
\begin{equation}
\begin{aligned}
\quad \ 1-\alpha^k
&=1-\sum_{i=1}^{N_k} \prod_{j=1}^{i-1}(1-\alpha_j^k)\cdot \alpha_i^k\\
&=1 - \alpha_1^k - (1-\alpha_1^k)\alpha_2^k - (1-\alpha_1^k)(1-\alpha_2^k)\alpha_3^k- \cdots\\
&= (1 - \alpha_1^k)(1 - \alpha_2^k - (1-\alpha_2^k)\alpha_3^k - \cdots ) \\
&=(1-\alpha_1^k)(1-\alpha_2^k)(1-\alpha_3^k - \cdots)\\
&= \cdots \\
&=\prod_{i=1}^{N_k}(1-\alpha_i^k)
\end{aligned}
\label{eq:1-a}
\end{equation}
Consequently, we can obtain the following by substituting~\cref{eq:1-a} into~\cref{eq:proof}. 


\begin{equation}
\begin{aligned}
\boldsymbol{C}_d(\boldsymbol{r})&=\sum_{k=1}^K\prod_{j=1}^{k-1}(1-\alpha^j)\cdot \boldsymbol{C}^k_d\\
&=\sum_{k=1}^K\prod_{j=1}^{k-1}\prod^{N_j}_{i=1}(1-\alpha_i^j)\cdot \boldsymbol{C}_d^k \\
&=\sum_{k=1}^K\prod^{N_1+\cdots+N_{k-1}}_{i=1}(1-\alpha_i)\cdot  \boldsymbol{C}_d^k
\end{aligned}
\label{eq:rendering result1}
\end{equation}
By substituting 
$\boldsymbol{C}_d^k$ from~\cref{eq:volume rendering inside blocks} into~\cref{eq:rendering result1}, we demonstrate the equivalence between~\cref{eq:tradition} and~\cref{eq:proof}.

\begin{align}
\boldsymbol{C}_d(\boldsymbol{r})
&=\sum_{k=1}^K\prod^{N_1+\cdots+N_{k-1}}_{l=1}(1-\alpha_l)\sum_{i=1}^{N_k}\prod_{j=1}^{i-1}(1-\alpha_j^k)\cdot\alpha_i^k\boldsymbol{c}_i^k\notag\\
&=\underbrace{\alpha_{1}^{1}\boldsymbol{c}_{1}^{1}+\cdots+ \prod_{i=1}^{N_{1}-1}(1-\alpha_i^1)\alpha_{N_{1}}^{1}\boldsymbol{c}_{N_{1}}^{1}}_{block 1}\notag\\
&+\underbrace{\prod_{j=1}^{N_{1}}(1-\alpha_j^1)\left( \alpha_{1}^{2}\boldsymbol{c}_{1}^{2}+\cdots+ \prod_{i=1}^{N_{2}-1}(1-\alpha_i^2)\alpha_{N_{2}}^{2}\boldsymbol{c}_{N_{2}}^{2}\right)}_{block 2} + \cdots \notag\\
&+ \underbrace{\prod_{j=1}^{N_{1}}(1-\alpha_j^1)\cdots \prod_{j=1}^{N_{K-1}}(1-\alpha_j^{K-1})\left( \alpha_{1}^{K}\boldsymbol{c}_{1}^{K}+\cdots+ \prod_{i=1}^{N_{K}-1}(1-\alpha_i^K)\alpha_{N_{K}}^{K}\boldsymbol{c}_{N_{K}}^{K}\right)}_{block K} \notag\\
&=\underbrace{\alpha_{1}^{1}\boldsymbol{c}_{1}^{1}+\cdots+ \prod_{i=1}^{N_{1}-1}(1-\alpha_i^1)\alpha_{N_{1}}^{1}\boldsymbol{c}_{N_{1}}^{1}}_{block 1}\\
&+\underbrace{\prod_{j=1}^{N_{1}}(1-\alpha_j^1)\alpha_{1}^{2}\boldsymbol{c}_{1}^{2}+\cdots+ \prod_{j=1}^{N_{1}}(1-\alpha_j^1)\prod_{i=1}^{N_{2}-1}(1-\alpha_i^2)\alpha_{N_{2}}^{2}\boldsymbol{c}_{N_{2}}^{2}}_{block 2} + \cdots \notag\\
&+\underbrace{\prod_{i=1}^{N_{1}}(1-\alpha_i^1)\cdots \prod_{i=1}^{N_{K-1}}(1-\alpha_i^{K-1}) \alpha_{1}^{K}\boldsymbol{c}_{1}^{K}+\cdots}_{block K}\quad \quad \quad \quad \quad \quad \quad \quad \quad \quad \quad \notag\\
&+ \underbrace{\prod_{i=1}^{N_{1}}(1-\alpha_i^1)\cdots \prod_{i=1}^{N_{K-1}}(1-\alpha_i^{K-1})\prod_{i=1}^{N_{K}-1}(1-\alpha_i^K)\alpha_{N_{K}}^{K}\boldsymbol{c}_{N_{K}}^{K})}_{block K} \notag\\
&=\underbrace{\alpha_{1}\boldsymbol{c}_{1}+\cdots+ \prod_{i=1}^{N_{1}-1}(1-\alpha_i)\alpha_{N_{1}}\boldsymbol{c}_{N_{1}}}_{block 1}\notag\\
&+\underbrace{\prod_{j=1}^{N_{1}}(1-\alpha_j) \alpha_{N_{1}+1}\boldsymbol{c}_{N_{1}+1}+\cdots+ \prod_{i=1}^{N_{1}+N_{2}-1}(1-\alpha_i)\alpha_{N_{1}+N_{2}}\boldsymbol{c}_{N_{1}+N_{2}}}_{block 2} + \cdots \notag\\
&+ \underbrace{\prod_{j=1}^{N_{1}+\cdots+N_{K-1}}(1-\alpha_j)\alpha_{N_{1}+\cdots+N_{K-1}+1}\boldsymbol{c}_{N_{1}+\cdots+N_{K-1}+1}+\cdots}_{block K} \notag\\
&+\underbrace{\prod_{i=1}^{N_{1}+\cdots+N_{K}-1}(1-\alpha_i)\alpha_{N_{1}+\cdots+N_{K}}\boldsymbol{c}_{N_{1}+\cdots+N_{K}}}_{block K} \notag\\
&=\sum_{i=1}^{N_{1}+\cdots+N_{K}}\prod^{i-1}_{j=1}(1-\alpha_j)\alpha_i\boldsymbol{c}_i\notag\\ 
&= \sum_{i=1}^{M}\prod^{i-1}_{j=1}(1-\alpha_j)\alpha_i\boldsymbol{c}_i = \hat{\boldsymbol{C}}_d(\boldsymbol{r}) \notag \\
\end{align}




It can be observed that block-based volume rendering approach in the resource-independent environment, such as in our case where multiple shaders are used to render the entire scene, is equivalent to the MERF’s volume rendering method~\cite{reiser2023merf} without deferred rendering. 

\section{More Details on Experiments}
In this section, we provide more details on our custom \textit{Campus} dataset, implemention details and settings of comparison methods.
\subsection{Dataset}

\begin{figure*}[t]
    \centering
    \includegraphics[width=1.0 \linewidth]{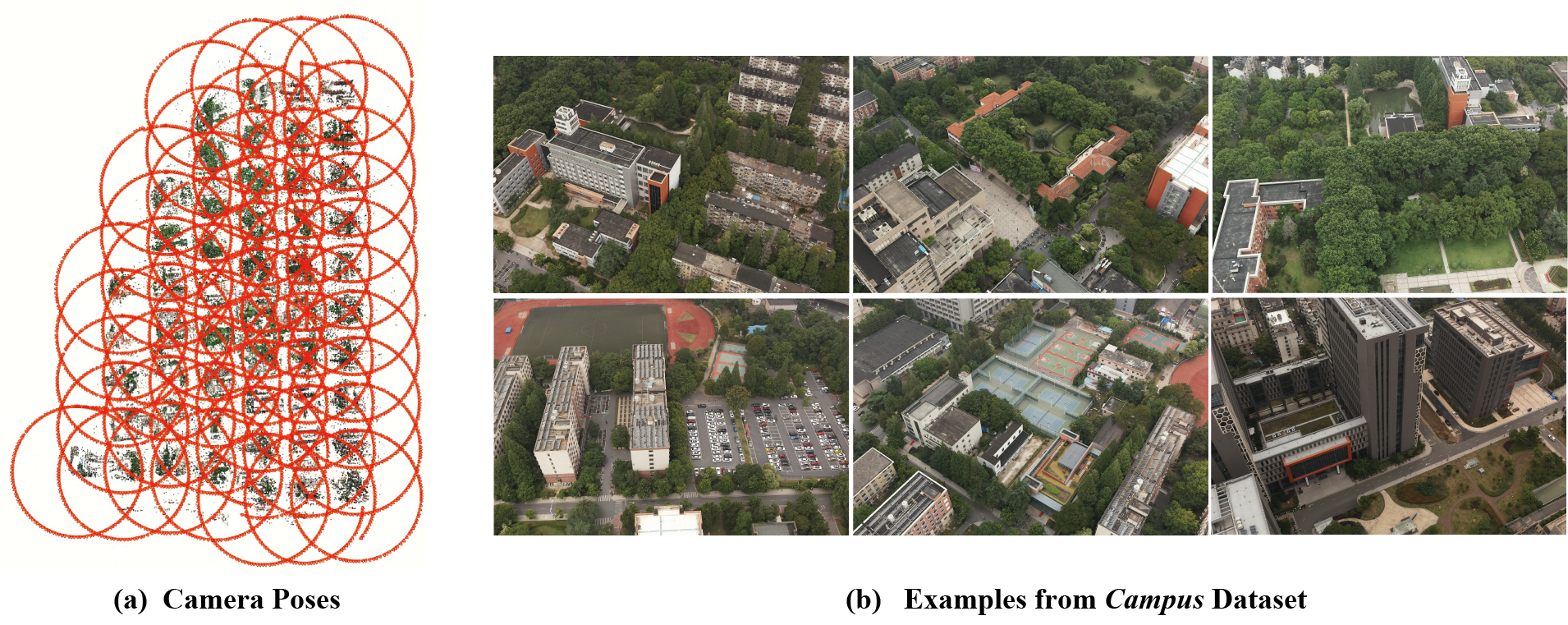}
    \caption{\textbf{Visulization of \textit{Campus} Dataset. }}
    \label{fig:cameras visualization}
\end{figure*}

The \textit{Campus} dataset is captured at an altitude of about 180 meters, covering an area of approximately 960,000 $m^2$. We adopt a circular data capture method for areial photography, as shown in \cref{fig:cameras visualization}. We find that this method often results in a higher overlap rate, allowing for a more accurate estimation of camera poses. Our dataset was captured over 8 hours on a cloudy day, with a fixed exposure setting to ensure almost identical appearance of photos taken at different time. We used Colmap~\cite{schoenberger2016sfm} to estimate camera poses. Feature matching was done using a vocabulary tree, followed by a hierarchical mapper followed by a few iterations of triangulation and bundle adjustment to estimate camera poses.

\subsection{Implementation Details}

For blocks at the boundaries of the entire scene, an unbounded scene representation is required to represent areas outside the block boundaries. We follow the same approach as MERF~\cite{reiser2023merf} to compute ray-AABB intersections trivially. To be specific, we employ the scene contraction function to project the scene external to the unit sphere into a cube, which has a radius of 2. The definition of the $j-th$ coordinate for a contracted point is as follows:

\begin{equation}
\operatorname{contract}(\mathbf{x})_j=\left\{\begin{array}{ll}
x_j & \text { if }\|\mathbf{x}\|_{\infty} \leq 1 \\
\frac{x_j}{\|\mathbf{x}\|_{\infty}} & \text { if } x_j \neq\|\mathbf{x}\|_{\infty}>1 \\
\left(2-\frac{1}{\left|x_j\right|}\right) \frac{x_j}{\left|x_j\right|} & \text { if } x_j=\|\mathbf{x}\|_{\infty}>1
\end{array},\right.
\end{equation}

We use an A100 GPU for training. In Sections 5.1 and 5.2, we perform training for 50,000 iterations with a batch size of 16384 pixels for the \textit{Campus} dataset, taking approximately 48 hours, while the training for other datasets takes around 24 hours. The experiments described in Sections 5.2 and 5.4 are subjected to training for 30,000 iterations with the same batch size, which is completed in approximately 12 hours. Training losses are initially balanced with $\lambda_{1} = 1.0$, $\lambda_{2} = 1.0$, $\lambda_{3} = 0.01$, $\lambda_{4} = 0.05$, $\lambda_{5} = 0.001$ and sample $2^{14}$ samples for computing sparsity loss.  During LOD genreation phase, we utilize the same loss function and hyperparameters as those employed during the training stage. We freeze the training of  submodels for 5,000 iterations and then refine them and shared global deferred mlp jointly for an additional 10,000 iterations.

\subsection{Comparative Method Settings}
In Sec. 5.1, we adopt the official implementations of Mega-NeRF~\cite{meganerf}, NeRFacto~\cite{tancik2023nerfstudio}, and Grid-NeRF~\cite{gridnerf}. Additionally, we use an unofficial implementation of Instant-NGP\footnote{https://github.com/ashawkey/torch-ngp}. Specifically, NeRFacto is trained with a batch size of 65,536 for 30,000 iterations. Instant-NGP~\cite{muller2022ngp} is trained for 500,000 iterations with a batch size of 4096. Grid-NeRF~\cite{gridnerf} is trained for 50,000 iterations with a batch size of 16384. Mega-NeRF~\cite{meganerf} is trained for 500,000 iterations for each block, using a batch size of 1024. In Sec. 5.2, For MobileNeRF~\cite{chen2023mobilenerf}, We initialize a $192^3$ grid to generate polygonal meshes while adhering to default parameters for other setups. We use the open-source version\footnote{https://github.com/hugoycj/torch-bakedsdf} for BakedSDF~\cite{yariv2023bakedsdf}, conducting training in two phases: 20,000 and 50,000 iterations, respectively with a batch size of 16,384 and use $1024^3$ to extract meshes by marching cubes~\cite{lorensen1998marching}. We employ the official implementation of MERF~\cite{reiser2023merf} and 3D Gaussian Splatting~\cite{kerbl3Dgaussians}.  During the MERF~\cite{reiser2023merf} experiments, the data is divided into four parts for a fair comparison with our method, and each block is trained with 32,768 batch size for 30,000 iterations, using default parameter settings.  For 3D Gaussian splatting~\cite{kerbl3Dgaussians}, we demonstrate the results after training for 30,000 and 100,000 iterations using the default settings.

\section{More Results}


\begin{figure*}[t]
    \centering
    \includegraphics[width=0.95 \linewidth]{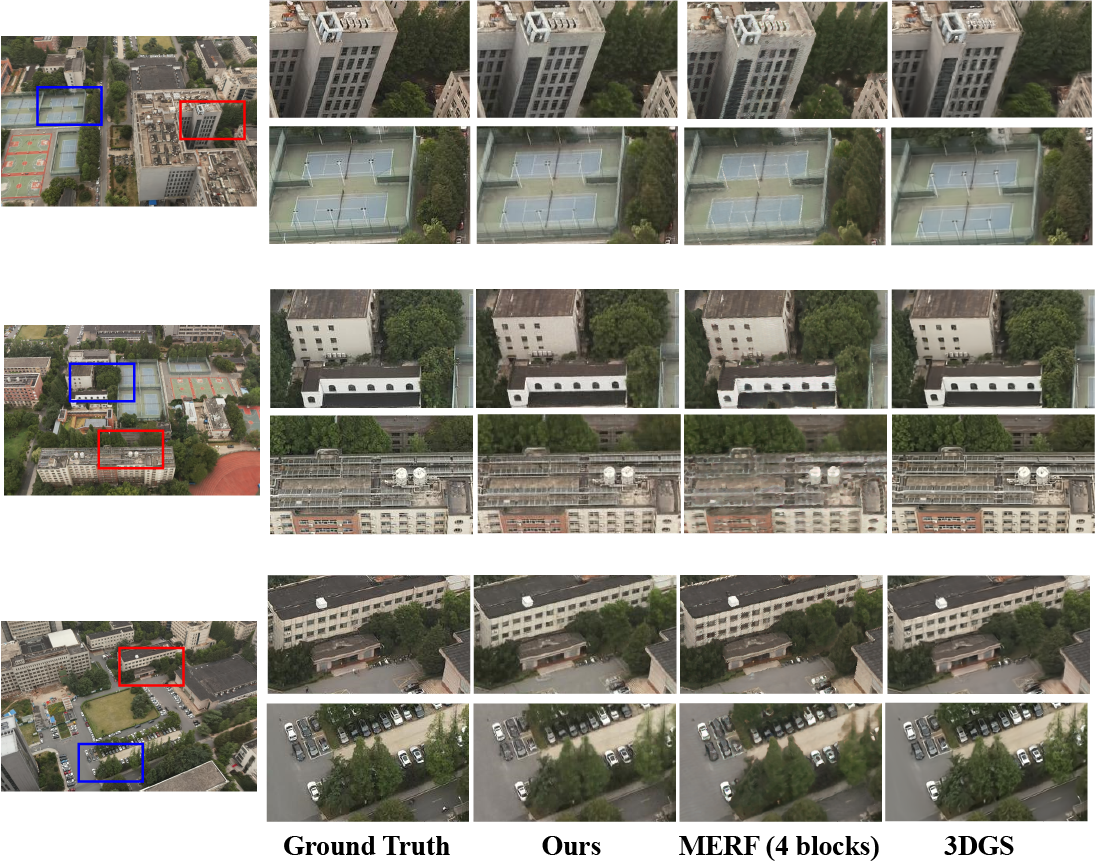}
    \caption{\textbf{Qualitative comparison of real-time rendering methods.}}
    \label{fig:more_real}
\end{figure*}  

\begin{figure*}
    \centering
    \includegraphics[width=0.95 \linewidth]{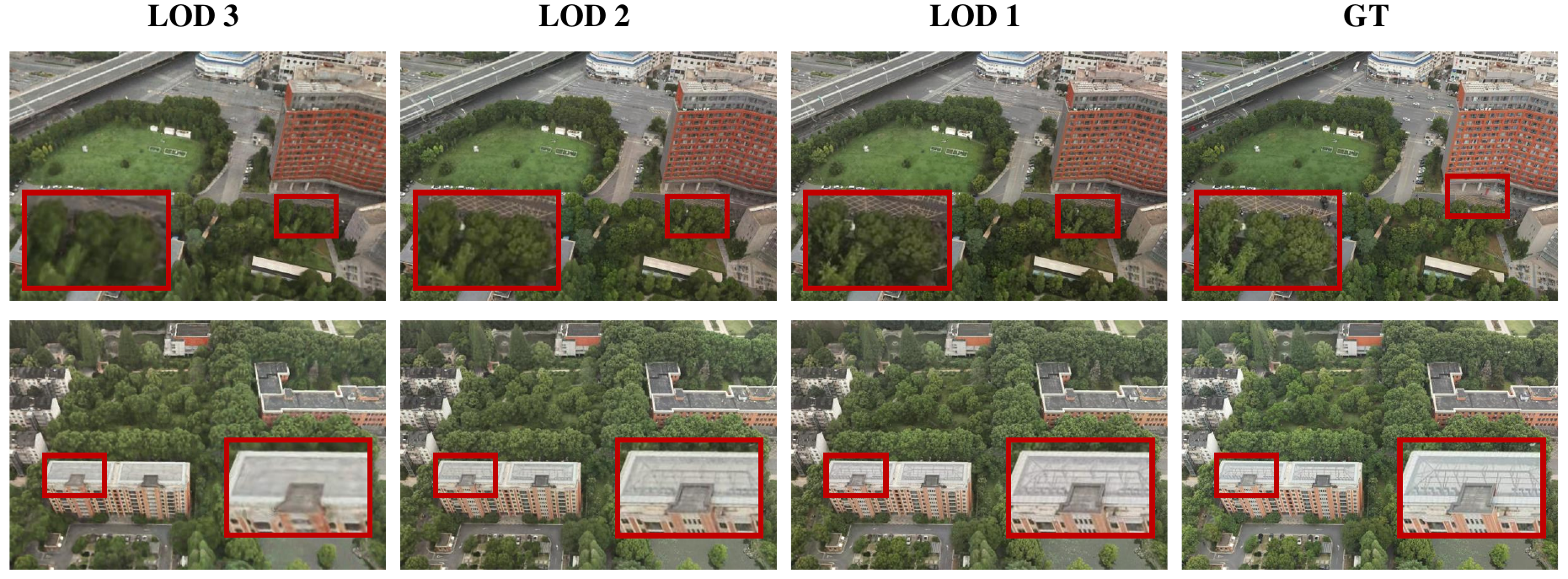}
    \caption{\textbf{Qualitative comparison of different levels-of-detail.}}
    \label{fig:more_lodl}
\end{figure*}

We present more qualitative comparisons among our method, MERF~\cite{reiser2023merf} (4 blocks), and 3D Gaussian Splatting~\cite{kerbl3Dgaussians} (100k iters) on the \textit{Campus} Dataset in~\cref{fig:more_real}.    Additionally, we showcase rendering results from different LODs to demonstrate the appearance consistency across various LODs in~\cref{fig:more_lodl}.

\section{Real-Time Viewer}
Our real-time viewer platform is based on the MERF volume renderer~\cite{reiser2023merf}, leveraging an OpenGL fragment shader to execute ray marching and deferred rendering. It accesses feature and density information through texture look-ups. However, we have implemented several improvements tailored for large scenes to enhance performance.

\textbf{Dynamic Loading.} We employ dynamic loading strategy to determine the  level-of-detail and  blocks to be loaded, thereby reducing VRAM usage. We use the center points obtained in Section 4.1 for each block on the $xy$ plane as the $xy$-components, and assign the height of the highest part within the block as the $z$-component. This defines the 3D center point for each block during the rendering stage. Additionally, we use the rectangle formed by the block's four $xy$ corner points and the previously determined $z$ value to define the block's region. By projecting this region onto the camera plane, we can determine whether the block is visible to the camera.

During the rendering phase, we first eliminate blocks that are not visible for cameras. Starting with the coarsest LOD, we check if the distance from the camera to all visible blocks within the finer LOD exceeds a certain threshold. For blocks beyond this threshold, rendering is performed using the resources loaded at the current LOD. Otherwise, the evaluation progresses to the subsequent, finer LOD. This stepwise process continues until we reach the finest LOD or complete the rendering of the entire image.

\textbf{Depth sorting.} We compute the distance from each block's center point to the camera using only their $xy$-components, leveraging this calculated distance for depth sorting. We adopt this approach as our scene segmentation occurs solely in the $xy$ plane, without division along the $z$-axis.

\section{Discussion}
\begin{wrapfigure}{r}{0.6\columnwidth}
    \vspace{-2.5em}
    \centering
    \includegraphics[width=1.0 \linewidth]{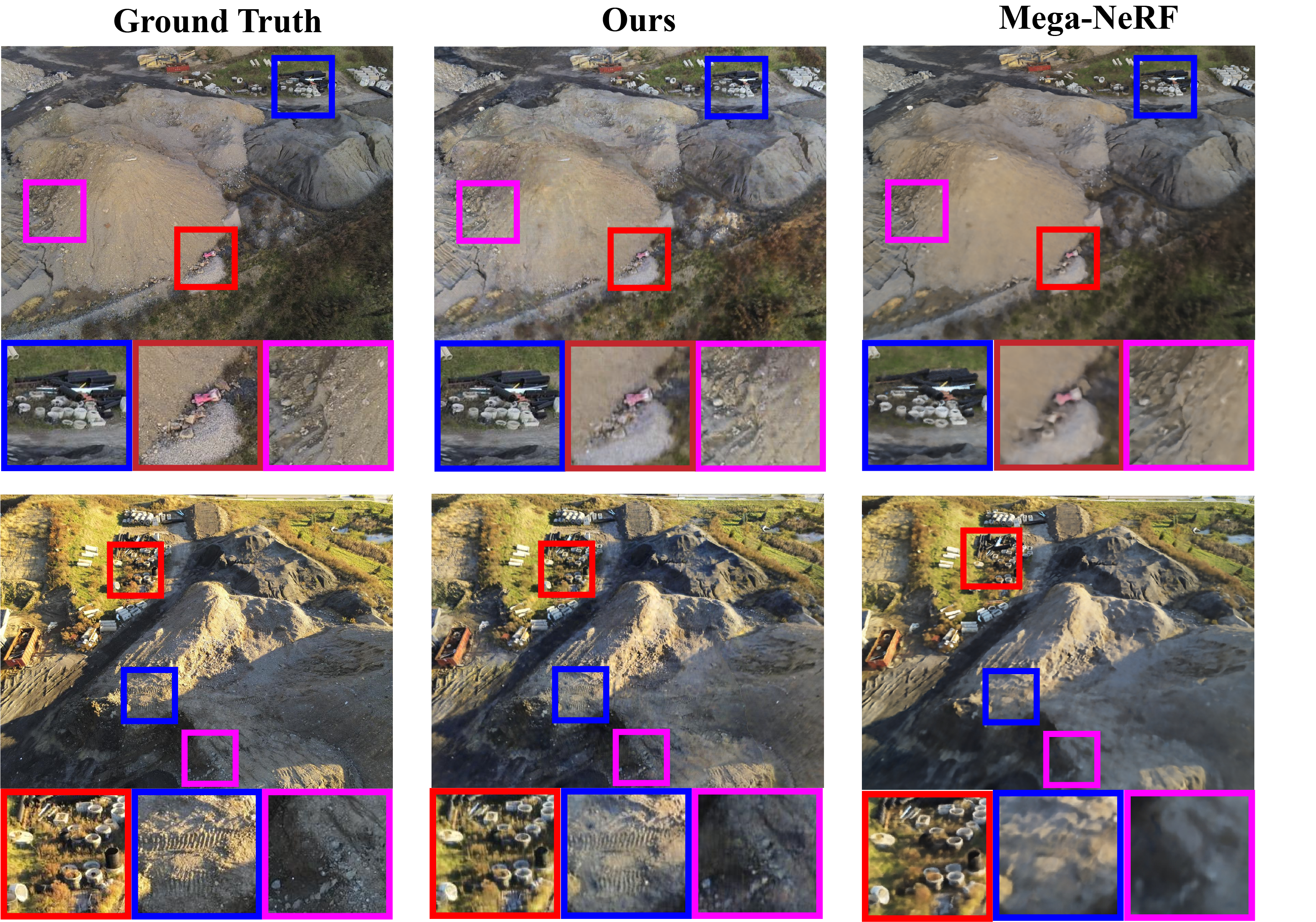}
    \caption{\textbf{Comparison to Mega-NeRF in\textit{ Rubble} dataset.} The \textit{Rubble} dataset exhibits significant lighting variations. The two images displayed above capture the same location but under vastly different lighting conditions. While our method reconstruct more details than Mega-NeRF~\cite{meganerf}, the deferred shading model has a limited capacity to represent the lighting and exposure variations as strong view-dependent effects. This limitations leads to marginally lower PSNR values.}
    \vspace{-3.0em}
  \label{fig:lighting}
\end{wrapfigure}
\textbf{Discussion.}  UE4-NeRF~\cite{gu2024ue4} and NeuRas~\cite{liu2023neural}, both mesh-based rendering approaches, offer fast rendering speed but face challenges when representing large, detail-rich scenes like those including dense foliage. These mesh representations can consume extensive memory, often amounting to several gigabytes.  Additionally, 3D Gaussian Splatting~\cite{kerbl3Dgaussians}, typically requires millions to tens of millions of points. The inherent properties of Gaussian splatting, which involve numerous Gauss-related attributes, further increase the VRAM needed for rendering. The substantial VRAM consumption characteristic of these methods poses significant challenges for implementing rendering on web platforms and resource-constrained devices, as they struggle to accommodate the high memory demands.

\textbf{Limitation \& Future Work.} Our method still has some limitations. Since we derive alpha blending across shaders based on the Lambertian surface assumption, visible seams may occur at the boundaries between blocks on non-Lambertian surfaces, such as water surfaces. Combining physically-based rendering with multiple shaders blending may alleviate this problem.  Additionally, while our approach achieves real-time rendering of large scenes on consumer-grade laptops, the inherently resource-intensive nature of large scenes makes that real-time rendering on mobile devices remains a challenge. Moreover, while our method recovers more intricate geometrical detail, it frequently results in color discrepancies with the ground truth image due to unstable lighting conditions and variable exposure, as shown in~\cref{fig:lighting}.

\end{document}